# An Exploration of Active Learning for Affective Digital Phenotyping

Peter Washington[1], MS; Cezmi Mutlu[4], MS; Aaron Kline[2], BS, MS; Cathy Hou[3]; Kaitlyn Dunlap[2], MS; Jack Kent[2], BA; Arman Husic[2], BS; Nate Stockham[5], MS; Brianna Chrisman[1], MS; Kelley Paskov[6], MS; Jae-Yoon Jung[2], PhD; PhD; Dennis P. Wall[2,6,*], PhD

[1]Department of Bioengineering
Stanford University
443 Via Ortega, Stanford, CA 94305

[2]Department of Pediatrics (Systems Medicine)
Stanford University
1265 Welch Rd, Stanford, CA 94305

[3]Department of Computer Science
Stanford University
353 Jane Stanford Way, Stanford, CA 94305

[4]Department of Electrical Engineering
Stanford University
350 Jane Stanford Way, Stanford, CA 94305

[5]Department of Neuroscience
Stanford University
213 Quarry Rd, Stanford, CA 94305

[6]Department of Biomedical Data Science
Stanford University
1265 Welch Rd Stanford, CA 94305

*To whom correspondence should be addressed: Dr. Dennis P. Wall, PhD
dpwall@stanford.edu
617-304-6031

# Abstract

Some of the most severe bottlenecks preventing widespread development of machine learning models for human behavior include a dearth of labeled training data and difficulty of acquiring high quality labels. Active learning is a paradigm for using algorithms to computationally select a useful subset of data points to label using metrics for model uncertainty and data similarity. We explore active learning for naturalistic computer vision emotion data, a particularly heterogeneous and complex data space due to inherently subjective labels. Using frames collected from gameplay acquired from a therapeutic smartphone game for children with autism, we run a simulation of active learning using gameplay prompts as metadata to aid in the active learning process. We find that active learning using information generated during gameplay slightly outperforms random selection of the same number of labeled frames. We next investigate a method to conduct active learning with subjective data, such as in affective computing, and where multiple crowdsourced labels can be acquired for each image. Using the Child Affective Facial Expression (CAFE) dataset, we simulate an active learning process for crowdsourcing many labels and find that prioritizing frames using the entropy of the crowdsourced label distribution results in lower categorical cross-entropy loss compared to random frame selection. Collectively, these results demonstrate pilot evaluations of two novel active learning approaches for subjective affective data collected in noisy settings.

# Introduction

Interactive mobile systems often generate large datasets of naturalistic behavior as a byproduct of their use [20-24], and these datasets can be leveraged to train new machine learning classifiers that may be more effective in real-time interactive systems. However, the data must first be labeled to be useful for machine learning. While data annotators could label the entirety of the dataset, this process can be infeasible for large and complex datasets. Neural networks take a long time to train - on the scale of days or weeks on the most powerful graphics processing units (GPUs) with large image datasets. This can result in thousands of dollars spent until training convergence. Active learning is a framework for identifying the most salient data points for humans to label in vast unlabeled datasets. Effective active learning strategies can drastically reduce the inefficiencies associated with labeling an entire dataset. An additional benefit of active learning is the generation of more focused and lightweight datasets with predictive power approaching that of the entire pool of data.

Active learning techniques typically use a machine learning model trained on a smaller labeled dataset to identify unlabeled data points with a low classifier probability. Traditional approaches to active learning include uncertainty-based approaches, where data for which the classifier is probabilistically uncertain of the correct class are prioritized, as well as a representation-based approaches, where similar unlabeled data points (where similarity may be determined by unsupervised approaches) are prioritized [37-38, 57]. Early approaches to active learning for affective computing have demonstrated noticeable improvements. Ahmed et al. achieve 88% emotion detection accuracy for 7 emotions (happy, disgusted, sad, angry, surprised, fear, and neutral) when applying active learning compared to less than 73% when selecting frames at random using a custom dataset curated from webcam images [1]. Their active learning method consists of evaluating mini batches of images with a pretrained model and replacing labels with a confidence below a static threshold. Thiam et al. also follows this paradigm by pretraining a background model on a dataset of emotional events and asking the human annotator to label examples that are not well explained by this background model, finding that 14% of the original dataset identified with active learning contains 75% of the total emotional events [43]. Senechal et al. apply active learning to the labeling of facial action units in naturalistic videos collected from webcams [36] to create a computationally tractable radial basis function support vector machine classifier reaching performance above the previously published baseline on AM-FED, a dataset of naturalistic facial expressions [31].

The largest challenge in active learning research is that there is no one-size-fits-all approach for all datasets [37-38]. For example, maximum entropy active learning consistently outperforms random selection of labels on some datasets while performing drastically worse than random on others [18, 35]. We have found that this issue is magnified when attempting to apply active learning on highly heterogeneous affective computer vision data.

Here, we explore active learning for affective datasets collected from an at-home digital phenotyping application for children with autism. Digital therapeutics for pediatric autism which help children practice emotion evocation [2, 5-7, 12-14, 26, 45-48, 61-62] are increasing being developed and studied for at-home therapy for use by families without access to traditional healthcare services. The use of these therapies generates massive datasets which contain emotion evocations. Because the therapies are administered by parents in home settings, the datasets are noisy and heterogeneous.

The dataset we use here consists of frames extracted from video recorded by a mobile autism therapeutic application consisting of a Charades game played by parents and children called *GuessWhat* [19, 20-24, 34, 49]. During game sessions, the parent places the phone on their forehead, facing the child. The child acts out the emotion prompt depicted on the device and the parent guesses the emotion performed by the child. If the parent tilts the phone forward, indicating a correct evocation of the emotion prompt by the child, then the corresponding frames receive the prompt as a metadata label. In the meantime, the front-facing camera of the phone records real time video of the interaction. We consider these "automatic label" metadata as potentially useful to the active learning process.

We run a post-hoc simulation of active learning on the dataset acquired from *GuessWhat* gameplay by ensuring that all emotions, as represented by the automatic label metadata, are equally represented. We try this approach with and without maximum entropy. Because we found marginal performance gains over random frame selection, we sought to understand why active learning was not more effective. We explore simulations with synthetic low-dimensional datasets to understand which dataset and model configurations fail during active learning procedures using noisy automatic labels. Finally, we recognize that emotion datasets often have subjective labels [50, 56], including in the *GuessWhat* dataset, so we pilot a crowdsourcing-based active learning algorithm to determine how many labels each data point should receive. This work provides preliminary explorations of the challenges and opportunities which arise in active learning with heterogeneous image datasets with noisy labels.

**Methods**

*Experiment 1: Active learning simulation using metadata from a mobile autism therapeutic*

We experiment with a dataset generated from *GuessWhat*, a smartphone app which provides at-home therapy to children with autism [20-24, 34]. During therapy sessions consisting of gameplay between a parent and child with autism the parent holds the smartphone on their forehead while the phone displays emotional Charades prompts to the child. If the child correctly evokes the emotion, then the parent tilts the phone forward, and this response is recorded using accelerometer sensors. The parent tilts the phone backwards if the child cannot correctly evoke the emotion. Throughout the gameplay session, *GuessWhat* collects video recordings of the child through the front-facing camera of the smartphone.

The gameplay sessions generate a video with timestamp labels marking the start and end frames of the prompt (Figure 1). Each frame is marked with the corresponding prompt on the *GuessWhat* session if any is present. These emotive metadata can be used as noisy automatic labels for training an emotion detection classifier. Here, we experiment with the potential of these metadata to aid in the active learning process.

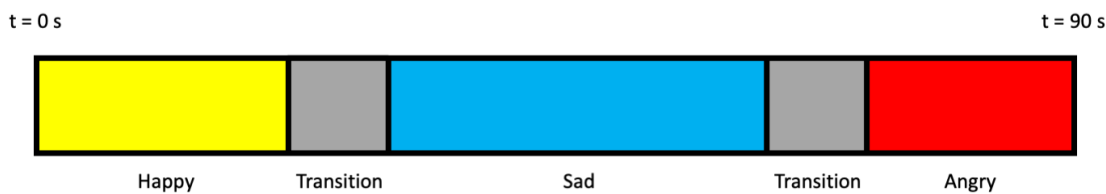

**Figure 1**. Generation of automatically labeled frames from gameplay. All time points during a 90 second gameplay session corresponds to either an emotion prompt or a transition in which the parent holding the phone is switching to the next emotion prompt.

After acquiring ground truth labels for all frames as described in Washington et al. [49], we ran a post hoc simulation using our final labeled dataset to understand the most effective strategy for ordering the frames to converge test performance with as few manual labels as possible (Figure 2). We evaluate the following active learning strategies, where $X$ is the set of unlabeled data frames and $Y$ is the set of possible predictions:

- **Baseline: random frames.** As a baseline without active learning, we order frames randomly. Sequential frames corresponding to a random emotion prompt from a random video session for a random child are consecutively displayed.

- **Automatic labels: cycle through (emotion, child) tuples.** There is a drastic class imbalance in the emotions that children evoked during gameplay sessions. To balance the labeled data to the extent possible, we cycle through one random frame for each (automatic emotion label, child) tuple until all tuples have been exhausted. The order of cycling through (automatic emotion label, child) tuples is chosen randomly.

- **Confidence-based approach using automatic labels: cycle through (emotion, child) tuples with maximum entropy frames first.** Same ordering as above, where (emotion, child) tuples are cycled through randomly, except instead of selecting a random (emotion, tuple) frame, the frame with the maximum entropy is selected. Entropy is defined as follows, where $x$ is the matrix of pixels constituting a frame, $y$ is the predicted class out of $N$ possible emotion classes, and $P(y|x)$ is the classifier's emitted probability of frame $x$ having class $y$:

$$H(x) = -\sum_{i=1}^{N} P(y_i|x)\log(P(y_i|x))$$

For each of these active learning strategies, we calculate a frame queue using that method and add the top 35 frames (5 per class) from the queue to the training set. We then train a classifier with the training set so far, recalculate the frame queue with the updated classifier, and repeat. We run 10 iterations of each active learning strategy.

We use a ResNet152V2 [17] convolutional neural network pretrained on ImageNet weights [8] for all active learning conditions and strategies. The network was trained with Adam optimization [25] and categorical cross-entropy loss. The following data augmentations were applied at random: rotation range of up to 7 degrees, a brightness range between 70% and 130%, and horizontal flips.

At each iteration, we evaluate the model on the Child Affective Facial Expression (CAFE) dataset [30], a publicly available image dataset of children emoting "anger", "disgust", "fear", "happy", "neutral", "sad", and "surprise" in controlled settings. CAFE contains 1,192 images of 90 females and 64 males. The racial distribution of subjects is: 27 African American, 16 Asian, 77 Caucasian/European American, 23 Latino, and 11 South Asian.

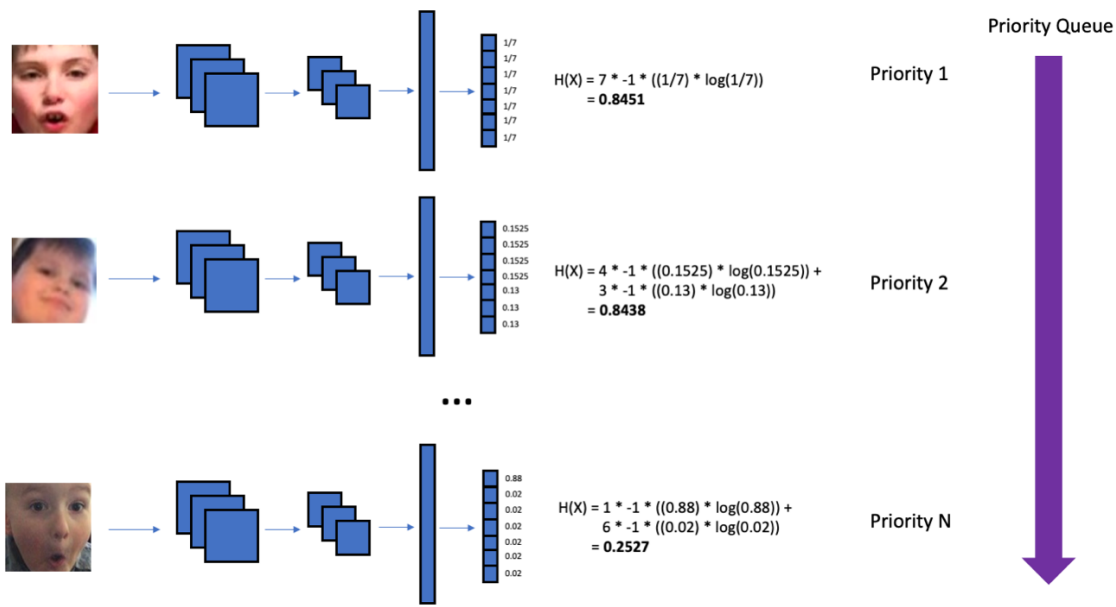

**Figure 2.** Maximum entropy-based active learning.

To emphasize the importance of working towards creating machine learning models that minimize bias and work for all individuals, we also evaluate the model against a subset of CAFE balanced by race, gender, and emotion expression. This subset contains N=125 images and has been used in prior literature due to its demographic balance [32].

Finally, we estimate how many frames are needed for the classifier to converge by fitting a power law distribution to the plot of classifier accuracy and F1 score vs. number of frames labeled. We use the *scipy* Python library to estimate parameters *a, b,* and *c* of the following equation, where *x* is the number of frames collected and *f(x)* is the accuracy or F1 score of the resulting classifier:

$$f(x) = (1 - a) - b\,x^c$$

*Experiment 2: Active learning with soft target labels*

Affective computing datasets often contain subjective labels. One case of subjective labels is for compound emotions. For example, a subject may be depicting compound emotions such as "happily surprised", "angrily surprised", or "fearfully surprised" [10]. In other cases, the label may be ambiguous. For example, some human raters may rate an image as depicting anger while others may rate it as disgusted [50, 56]. Both answers could be correct, or the true expression might require additional context. To account for this type of subjectivity in emotional labels, training emotion classifiers with soft-target labels rather than one-hot encoded labels has been proposed and validated as a feasible method to create classifiers which output probability distributions which mimic the variation in human interpretation [50, 56].

We investigate active learning with such subjective data where multiple labels are required per image to construct a soft-targe label from the crowd. We use CAFE for an active learning simulation, as CAFE contains 100 independent human annotations per image. We evaluate on 3 separate folds of CAFE to verify robustness of results. Each of the 3 folds consists of two-thirds of CAFE in the training set and the rest in the test set. To avoid overfitting, we ensured that no child subject in the training set ever appeared in the test set for all folds. The final one-hot labels were determined by majority voting, with ties broken randomly.

At each iteration of the simulation, we randomly sample labels from the distribution of 100 human annotations provided in CAFE for all images in the CAFE dataset. A total of 3N labels are sampled per iteration, where N is the total number of images. In the baseline condition, we sample 3 labels per image. We train a ResNet classifier for 200 epochs with the same configuration as the ResNet in Experiment 1. To generate each number at each iteration, we take the mean and standard deviation of the held-out validation data for the final 50 epochs. Training accuracy converged prior to epoch 150 for all conditions, folds, and simulation iterations.

In the active learning condition, we sample 1 label for all images. For the remainder of labels, we sort the remaining frames by decreasing entropy of the crowd label distribution. For the frames in the top 10[th] percentile of maximum entropy crowd labels, we sample 6 additional times (7 samples total). For the frames in the 10[th] through 25[th] percentile, we sample 4 additional times (5 samples total). For frames in the 25[th] through 65[th] percentile, we sample 2 additional times (3 samples total). The remaining frames do not receive additional labels at that iteration (1 sample total). This again results in 3N samples distributed in total in the active learning condition (N + 6*0.1*N + 4*0.15*N + 2*0.4*N = 3N).

**Results**

*Experiment 1: Active learning simulation using data from a mobile autism therapeutic*

We first perform a simulation without including any public datasets, as training time to convergence would drastically increase with the additional number of training frames. We first evaluate against a subset of CAFE balanced by emotion, gender, and race, which we call the "balanced CAFE subset". This balanced CAFE subset has been evaluated in prior work. [32]. The network trained on all images achieved an overall accuracy of 84.8% accuracy and an 83.9% F1-score on the balanced CAFE subset. When excluding the Hollywood Squares dataset and only training using the RaFD, CK+, and JAFFE datasets, the resulting model achieves 70.4% accuracy on the balanced CAFE subset, highlighting the added value of the *GuessWhat* data.

Cycling through (user-derived emotion label, child) tuples and selecting the highest entropy frame satisfying that tuple outperforms selection of random frames by a large margin and outperforms cycling through tuples and choosing a random frame in that tuple by a smaller margin (Figure 3). Cycling through frame metadata with maximum entropy starts to reach near convergence at 4 iterations (140 frames) while cycling through frame metadata alone does no reach this point until at least iteration 8 (305 frames).

(A)

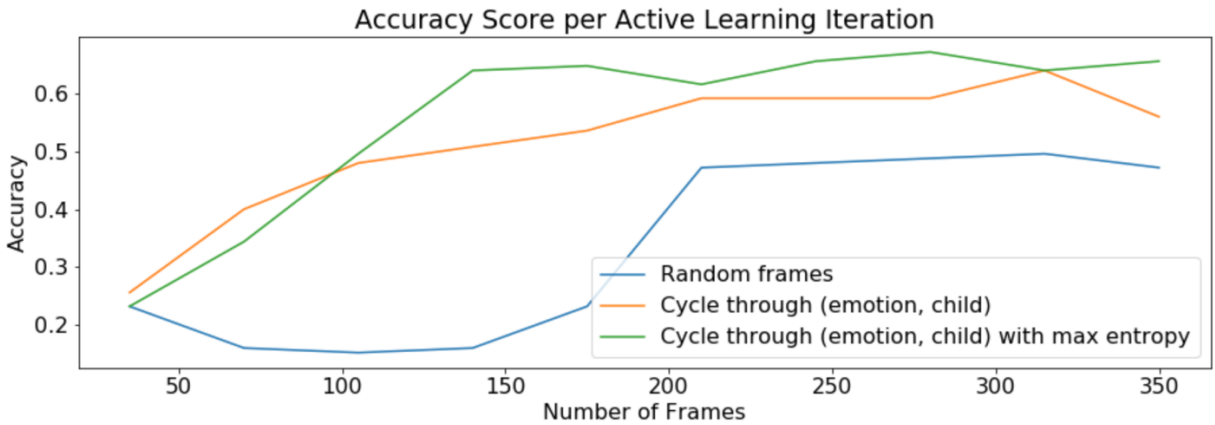

(B)

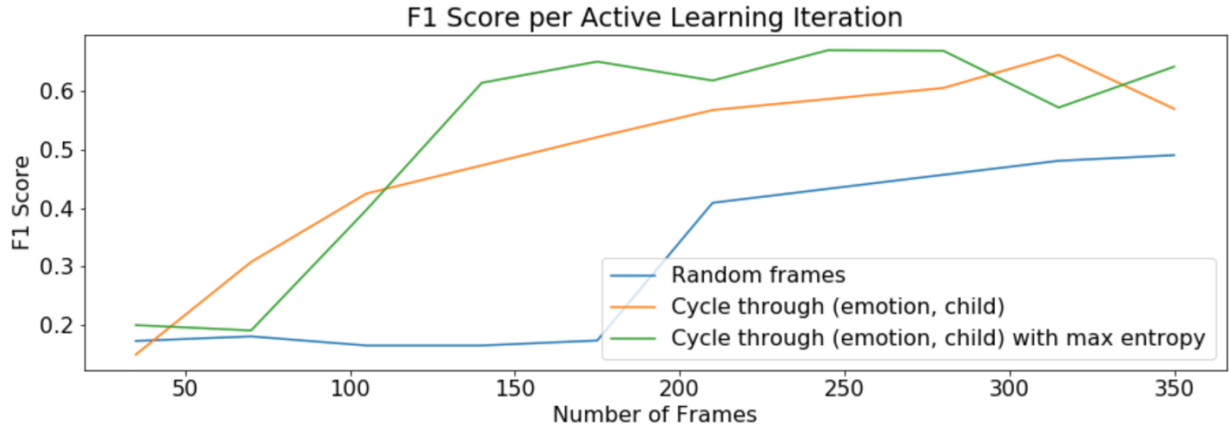

**Figure 3**. Comparison of active learning techniques to baseline when evaluating against a balanced subset of CAFE images. The x-axis represents the total number of frames included in the dataset. Cycling through (user-derived emotion label, child) tuples and selecting the highest entropy frame satisfying that tuple (green) is compared against cycling through tuples and choosing a random frame in that tuple (orange) as well as only selecting random frames (blue). 35 frames, or 5 frames per class, are chosen per active learning iteration. (A) Accuracy per active learning iteration. (B) F1-score per active learning iteration.

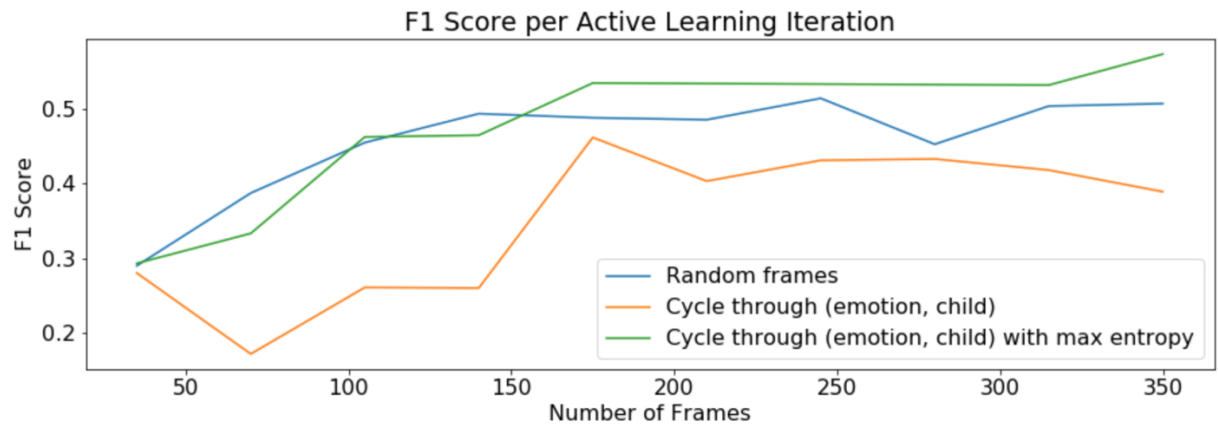

**Figure 4**. F1 score per active learning iteration when evaluating against the entirety of CAFE. The x-axis represents the total number of frames included in the dataset.

We next evaluate active learning on the entirety of CAFE. Figure 4 shows that cycling through (emotion autolabel, child) tuples and selecting the maximum entropy frame outperforms random baseline when evaluating on the entirety of CAFE, although the margin is not as dramatic as compared to the balanced CAFE subset. Interestingly, cycling through (emotion autolabel, child) tuples without choosing the maximum entropy frame performs worse than the random baseline. This is possibly due to the severe class imbalance of the full CAFE dataset. Figure 5 shows that cycling through (emotion, child) tuples and selecting the maximum entropy frame also outperforms random baseline when starting with a model pretrained on JAFFE and CK+.

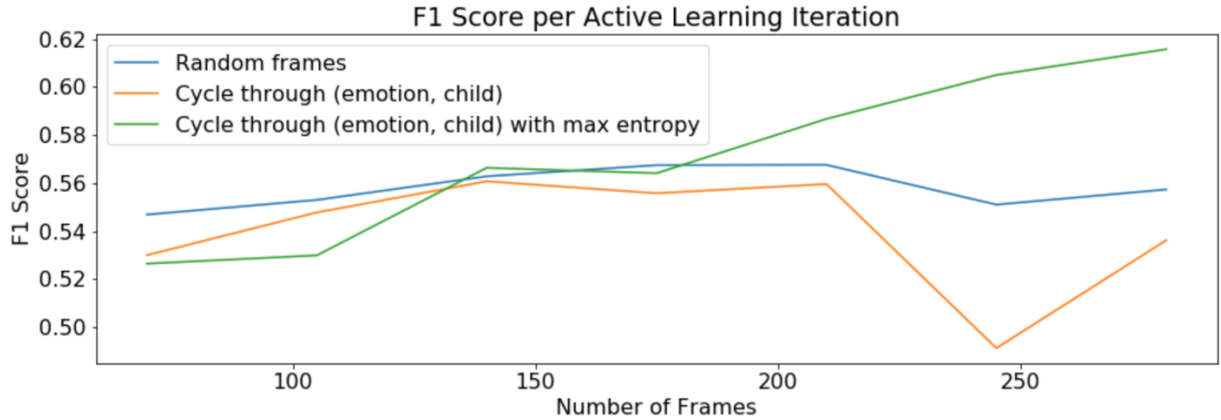

**Figure 5.** F1 score per active learning iteration when evaluating against the entirety of CAFE. The x-axis represents the total number of additional frames added to the dataset for a baseline consisting of the entirety of the JAFFE and CK+ datasets.

Figure 6 shows an active learning simulation when including JAFFE and CK+ prior to the first active learning iteration and evaluated on the balanced CAFE subset. Fitting a power law distribution to the data, we arrive at the following relationship:

$$CAFE\ Accuuracy = 0.704 + 0.008\ (number\ of\ labeled\ frames)^{0.257}$$

This relationship predicts 81.49% accuracy with 35,265 labeled frames, which is a slight underprediction of the true accuracy of 84.8% when training a ResNet with all frames. This is likely a result of the decreased utility of progressively low entropy frames.

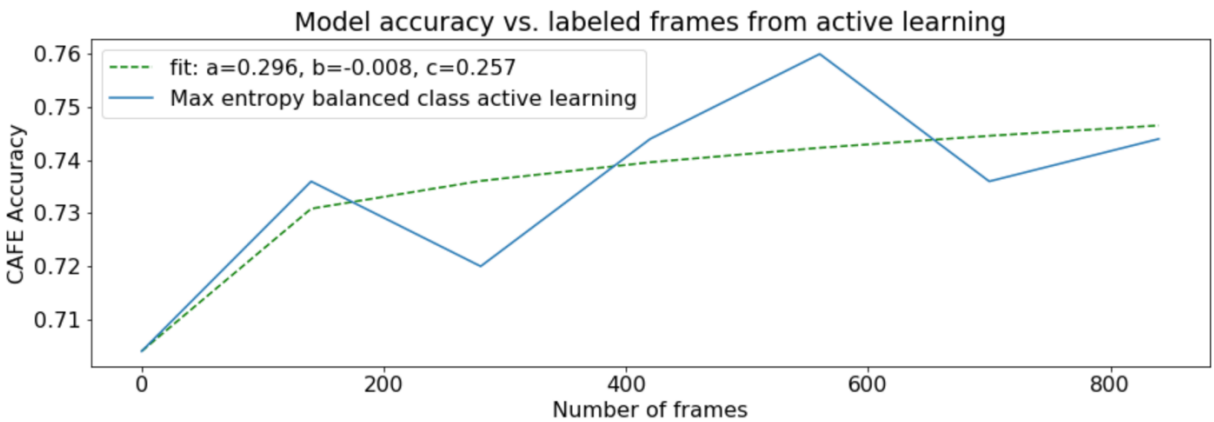

**Figure 6.** Fitting a power law distribution to the plot of balanced CAFE subset accuracy vs. the number of frames labeled using active learning when including JAFFE, RaFD, and CK+ prior to the first active learning iteration.

*Experiment 2: Active learning with soft targets*

We recorded the categorical cross-entropy loss when training and testing with one-hot encoded labels (Table 1), training and testing with soft-target labels (Table S1), training with one-hot encoded labels and testing with soft-target labels (Table S2), and training with soft-target labels and testing with one-hot encoded labels (Table S3). We find that active learning with crowd labels consistently outperforms random cycling of frames when training and testing with one-hot encoded labels across all folds: in all 11 iterations for fold 1, in 9 out of 11 iterations in fold 2, and in 8 out of 11 iterations in fold 3.

| Labeled Frames | *Fold 1* | | *Fold 2* | | *Fold 3* | |
|---|---|---|---|---|---|---|
| | Cycling | Active Learning | Cycling | Active Learning | Cycling | Active Learning |
| 3 | 1.56 ± 0.26 | 1.47 ± 0.33 | 1.84 ± 0.35 | 1.83 ± 0.37 | <span style="color:red">1.69 ± 0.29</span> | <span style="color:red">1.88 ± 0.43</span> |
| 6 | 1.49 ± 0.35 | 1.42 ± 0.26 | 2.05 ± 0.59 | 1.98 ± 0.48 | 2.02 ± 0.40 | 1.86 ± 0.35 |
| 9 | 1.75 ± 0.49 | 1.47 ± 0.31 | <span style="color:red">1.99 ± 0.56</span> | <span style="color:red">2.02 ± 0.43</span> | 2.03 ± 0.45 | 2.00 ± 0.51 |
| 12 | 1.60 ± 0.36 | 1.55 ± 0.28 | 2.09 ± 0.56 | 1.89 ± 0.47 | 2.04 ± 0.74 | 2.03 ± 0.44 |
| 15 | 1.60 ± 0.38 | 1.59 ± 0.38 | <span style="color:red">2.08 ± 0.38</span> | <span style="color:red">2.24 ± 0.47</span> | 2.24 ± 0.54 | 2.12 ± 0.67 |
| 18 | 1.74 ± 0.45 | 1.63 ± 0.37 | 2.05 ± 0.52 | 1.93 ± 0.39 | <span style="color:red">1.69 ± 0.36</span> | <span style="color:red">2.39 ± 0.60</span> |
| 21 | 1.70 ± 0.36 | 1.61 ± 0.34 | 2.08 ± 0.70 | 1.99 ± 0.53 | 2.26 ± 0.43 | 1.87 ± 0.40 |
| 24 | 1.62 ± 0.32 | 1.53 ± 0.34 | 2.50 ± 0.63 | 1.93 ± 0.56 | <span style="color:red">1.93 ± 0.39</span> | <span style="color:red">2.00 ± 0.44</span> |
| 30 | 1.90 ± 0.38 | 1.58 ± 0.32 | 2.11 ± 0.53 | 2.06 ± 0.49 | 1.94 ± 0.45 | 1.65 ± 0.37 |
| 45 | 1.56 ± 0.26 | 1.47 ± 0.33 | 2.34 ± 0.63 | 2.05 ± 0.49 | 2.03 ± 0.56 | 1.98 ± 0.47 |
| 75 | 1.76 ± 0.49 | 1.47 ± 0.31 | 2.60 ± 0.51 | 2.45 ± 0.61 | 2.13 ± 0.48 | 1.85 ± 0.33 |

Table 1. Categorical cross-entropy loss when training and testing with one-hot encoded labels after various iterations of frame labeling. The number of labeled frames is beyond the first 3 labels acquired per frame. Instances where the categorical cross-entropy loss is higher (by greater than 0.01) for active learning than equally labeling all frames are highlighted in red.

By contrast, active learning did not demonstrate any benefit over random cycling of frames when using soft-target labels in either the training or testing process. When training with soft-target labels (Tables S1 and S3), the losses between active learning and cycling conditions are nearly identical across all simulation iterations both when one-hot encoded and soft-target labels are used to evaluate the test set. When training with one-hot labels and testing on soft-targets (Table S2), there is no clear trend, with active learning performing better on some iterations and cycling performing better on others.

Using soft-target labels for emotion classification has resulted in lower loss compared to one-hot encoded labels in prior works [50, 56]. We replicate these results here: when testing with one-hot encoded labels, training with soft-target labels (Table S3) results in noticeably lower loss than training with one-hot encoded labels (Table 1) by a margin larger than the error bars. When testing with soft-target labels, training with soft-target labels (Table S1) results in similarly lower loss compared to training with one-hot encodings (Table S2).

**Discussion**

We explore some of the challenges associated with active learning for heterogeneous, noisy, and subjective datasets such as emotion data. The first approach we take incorporates metadata which are collected simultaneously with the original dataset. These metadata provide a noisy prior estimation of the true label, allowing for ensuring of some degree of class balance (with some degree of error) during the active learning process. We find that this approach outperforms random cycling of frames. We recommend further explorations of the utility of automatically generated metadata in machine learning processes.

While we scratch the surface of exploring the incorporation of metadata in the active learning process, there are several possible areas of further study. Modern active learning approaches have been proposed which work well for deep learning with image data. For example, Bayesian convolutional neural networks have been demonstrated to aid active learning for high dimensional image datasets [11], as the neural network weights are modeled as distributions and the output of the neural network can therefore be interpreted as a calibrated probability distribution.

The major limitation of this portion of the study is the exploration of this technique on only a single dataset. Further studies should validate this method on other datasets with different properties, including for classifiers related to other behavioral features of autism [3-4, 9, 15-16, 27-28, 33, 39, 44, 52]. One possible source of confounding variation in the automatic labels could be different biases towards certain incorrect classes. In methods such as the one presented here where the metadata plays a fundamental role in which data are prioritized, bias mitigation strategies for the metadata are equally as important as strategies for the data itself. Avoiding bias in active learning in general is an under-explored topic.

Another challenge of active learning with "difficult data" containing subjective classes is the incorporation of multiple independent human annotations in the active learning process. In particular, we study active learning in situations where multiple crowdsourced acquisitions are acquired for each video, a common practice in digital behavioral phenotyping [29, 40-42, 51, 53-60]. We find that prioritizing frames with the maximum entropy according to the distribution of crowdsourced responses results in lower loss than obtaining an equal number of human labels per image. This result does not hold for incorporating soft-target labels in either the training or test sets. These results indicate that crowdsourced active learning is useful as an uncertainty metric when crowdsourcing the labeling of subjective data points and training with standard approaches.

**Conclusion**

Active learning is challenging, as one-size-fits-all approaches do not work for all datasets, and the difficulty of creating useful active learning algorithms is exacerbated when the data are heterogeneous, noisy, and subjective. We propose two approaches to active learning with affective computing datasets in two situations: (1) when noisy metadata provide an estimation of the true class to enable balanced prioritization of data points, and (2) when multiple labels are acquired for each image.

**Acknowledgements**
The work was supported in part by funds to DPW from the National Institutes of Health (1R01EB025025-01, 1R01LM013364-01, 1R21HD091500-01, 1R01LM013083), the National Science Foundation (Award 2014232), The Hartwell Foundation, Bill and Melinda Gates Foundation, Coulter Foundation, Lucile Packard Foundation, Auxiliaries Endowment, the ISDB Transform Fund, the Weston Havens Foundation, and program grants from Stanford's Human Centered Artificial Intelligence Program, Precision Health and Integrated Diagnostics Center, Beckman Center, Bio-X Center, Predictives and Diagnostics Accelerator, Spectrum, Spark Program in Translational Research, MediaX, and from the Wu Tsai Neurosciences Institute's Neuroscience:Translate Program. We also acknowledge generous support from David Orr, Imma Calvo, Bobby Dekesyer and Peter Sullivan. PW would like to acknowledge support from Mr. Schroeder and the Stanford Interdisciplinary Graduate Fellowship (SIGF) as the Schroeder Family Goldman Sachs Graduate Fellow.

**Supplementary Material**

|  | *Fold 1* | | *Fold 2* | | *Fold 3* | |
|---|---|---|---|---|---|---|
| **Labeled Frames** | **Cycling** | **Active Learning** | **Cycling** | **Active Learning** | **Cycling** | **Active Learning** |
| 3 | 0.96 ± 0.01 | 0.97 ± 0.01 | 1.00 ± 0.01 | 0.99 ± 0.01 | 1.04 ± 0.02 | 1.06 ± 0.01 |
| 6 | 0.97 ± 0.01 | 0.96 ± 0.01 | 1.00 ± 0.01 | 1.00 ± 0.01 | 1.04 ± 0.01 | 1.04 ± 0.01 |
| 9 | 0.96 ± 0.01 | 0.96 ± 0.01 | 1.00 ± 0.01 | 0.99 ± 0.01 | 1.05 ± 0.01 | 1.05 ± 0.01 |
| 12 | 0.97 ± 0.01 | 0.96 ± 0.01 | 0.99 ± 0.01 | 0.99 ± 0.01 | 1.04 ± 0.01 | 1.03 ± 0.01 |
| 21 | 0.96 ± 0.01 | 0.97 ± 0.01 | 1.00 ± 0.01 | 0.99 ± 0.01 | 1.05 ± 0.01 | 1.05 ± 0.01 |
| 30 | 0.97 ± 0.01 | 0.96 ± 0.01 | 0.99 ± 0.01 | 1.00 ± 0.01 | 1.05 ± 0.01 | 1.04 ± 0.01 |
| 45 | 0.96 ± 0.01 | 0.96 ± 0.01 | 0.99 ± 0.01 | 1.00 ± 0.01 | 1.05 ± 0.01 | 1.05 ± 0.01 |
| 75 | 0.96 ± 0.01 | 0.96 ± 0.01 | 1.01 ± 0.01 | 0.99 ± 0.01 | 1.04 ± 0.01 | 1.04 ± 0.01 |

Table S1. Categorical cross-entropy loss when training and testing with soft-target labels after various iterations of frame labeling. The number of labeled frames is beyond the first 3 labels acquired per frame.

|  | *Fold 1* | | *Fold 2* | | *Fold 3* | |
|---|---|---|---|---|---|---|
| **Labeled Frames** | **Cycling** | **Active Learning** | **Cycling** | **Active Learning** | **Cycling** | **Active Learning** |
| 3 | 3.24 ± 0.51 | 2.75 ± 0.67 | 3.39 ± 0.49 | 3.20 ± 0.53 | 2.81 ± 0.61 | 3.04 ± 0.48 |
| 6 | 2.88 ± 0.56 | 3.13 ± 0.53 | 3.14 ± 0.68 | 3.37 ± 0.73 | 3.39 ± 0.62 | 3.22 ± 0.58 |
| 9 | 3.26 ± 0.71 | 3.06 ± 0.44 | 3.16 ± 0.80 | 3.49 ± 0.71 | 3.55 ± 0.76 | 3.46 ± 0.68 |
| 12 | 3.13 ± 0.47 | 3.04 ± 0.68 | 3.34 ± 0.60 | 3.25 ± 0.57 | 3.47 ± 0.88 | 3.51 ± 0.71 |
| 21 | 3.38 ± 0.42 | 3.09 ± 0.49 | 3.20 ± 0.74 | 3.62 ± 0.47 | 3.72 ± 0.64 | 3.37 ± 0.75 |
| 30 | 3.54 ± 0.46 | 3.13 ± 0.59 | 3.42 ± 0.57 | 3.41 ± 0.65 | 3.62 ± 0.51 | 3.28 ± 0.55 |
| 45 | 3.24 ± 0.51 | 2.75 ± 0.67 | 3.29 ± 0.76 | 3.46 ± 0.57 | 3.76 ± 0.81 | 3.47 ± 0.71 |
| 75 | 3.26 ± 0.71 | 3.06 ± 0.43 | 3.51 ± 0.49 | 3.63 ± 0.72 | 3.41 ± 0.65 | 3.19 ± 0.64 |

Table S2. Categorical cross-entropy loss when training with one-hot encoded labels and testing with soft-target labels after various iterations of frame labeling. The number of labeled frames is beyond the first 3 labels acquired per frame.

|                | Fold 1 | | Fold 2 | | Fold 3 | |
|----------------|--------|--------|--------|--------|--------|--------|
| Labeled Frames | Cycling | Active Learning | Cycling | Active Learning | Cycling | Active Learning |
| 3  | $0.60 \pm 0.02$ | $0.61 \pm 0.02$ | $0.69 \pm 0.03$ | $0.69 \pm 0.02$ | $0.71 \pm 0.03$ | $0.71 \pm 0.02$ |
| 6  | $0.62 \pm 0.02$ | $0.62 \pm 0.02$ | $0.70 \pm 0.03$ | $0.70 \pm 0.03$ | $0.70 \pm 0.03$ | $0.70 \pm 0.03$ |
| 9  | $0.61 \pm 0.02$ | $0.61 \pm 0.02$ | $0.70 \pm 0.03$ | $0.68 \pm 0.03$ | $0.71 \pm 0.02$ | $0.71 \pm 0.03$ |
| 12 | $0.62 \pm 0.02$ | $0.61 \pm 0.02$ | $0.70 \pm 0.03$ | $0.70 \pm 0.02$ | $0.70 \pm 0.02$ | $0.68 \pm 0.02$ |
| 21 | $0.61 \pm 0.02$ | $0.62 \pm 0.02$ | $0.68 \pm 0.03$ | $0.69 \pm 0.03$ | $0.73 \pm 0.03$ | $0.72 \pm 0.02$ |
| 30 | $0.62 \pm 0.02$ | $0.61 \pm 0.02$ | $0.69 \pm 0.03$ | $0.71 \pm 0.03$ | $0.69 \pm 0.02$ | $0.69 \pm 0.03$ |
| 45 | $0.60 \pm 0.02$ | $0.61 \pm 0.02$ | $0.72 \pm 0.02$ | $0.68 \pm 0.02$ | $0.70 \pm 0.02$ | $0.71 \pm 0.03$ |
| 75 | $0.61 \pm 0.02$ | $0.61 \pm 0.02$ | $0.71 \pm 0.03$ | $0.67 \pm 0.02$ | $0.70 \pm 0.02$ | $0.69 \pm 0.02$ |

Table S3. Categorical cross-entropy loss when training with soft-target labels and testing with one-hot encoded labels after various iterations of frame labeling. The number of labeled frames is beyond the first 3 labels acquired per frame.